\documentclass[12pt]{article}

\usepackage[utf8]{inputenc}
\usepackage[english]{babel}
\usepackage[T1]{fontenc}
\usepackage{multirow}
\usepackage{graphicx}
\usepackage{array}
\usepackage[linesnumbered,ruled,vlined]{algorithm2e}
\usepackage[a4paper, margin=2.7cm]{geometry}
\usepackage{graphicx}
\usepackage{subcaption}
\usepackage{enumerate}
\usepackage{amsmath, amsthm, amsfonts, amssymb}
\usepackage{mathrsfs}           
\usepackage[normalem]{ulem}
\useunder{\uline}{\ul}{}
\newcommand\tab[1][1cm]{\hspace*{#1}}

\usepackage[HTML]{xcolor}
\usepackage[export]{adjustbox}
\definecolor{MyColor}{HTML}{000000}

\usepackage{float}

\usepackage[colorlinks=true,linkcolor=blue,citecolor=blue,urlcolor=blue,breaklinks]{hyperref}

\SetKwInput{KwInput}{Input}                
\SetKwInput{KwOutput}{Output}              

\usepackage{bbm} 

\date{}

\begin{document}

\begin{flushleft}

\begin{center}

{\Large
\textbf{GuideWalk: A Novel Graph - Based Word Embedding for Enhanced Text Classification} \par
}

\bigskip

Sarmad N. MOHAMMED\textsuperscript{1,2*},
Semra G\"UND\"U\c{C}\textsuperscript{2}
\\
\bigskip

{ \footnotesize
$^{1}$  Computer Science Department, College of Computer Science and Information Technology, University of Kirkuk, Kirkuk, Iraq
\\
$^{2}$ Computer Engineering Department, Engineering Faculty, Ankara University, Ankara, Turkey

\bigskip

* sarmad\_mohammed@uokirkuk.edu.iq
}
\end{center}

\end{flushleft}

\begin{abstract}
One of the prime problems of computer science and machine learning is to extract information efficiently from large-scale, heterogeneous data. Text data, with its syntax, semantics, and even hidden information content, possesses an exceptional place among the data types in concern. The processing of the text data requires embedding, a method of translating the content of the text to numeric vectors. A correct embedding algorithm is the starting point for obtaining the full information content of the text data. In this work, a new text embedding approach, namely the Guided Transition Probability Matrix (GTPM) model is proposed. The model uses the graph structure of sentences to capture different types of information from text data, such as syntactic, semantic, and hidden content. Using random walks on a weighted word graph, GTPM calculates transition probabilities to derive text embedding vectors. The proposed method is tested with real-world data sets and eight well-known and successful embedding algorithms. GTPM shows significantly better classification performance for binary and multi-class datasets than well-known algorithms. Additionally, the proposed method demonstrates superior robustness, maintaining performance with limited (only $10\%$) training data, showing an $8\%$ decline compared to $15-20\%$ for baseline methods. 
  
  
\end{abstract}

\section{Introduction}
\label{intro}
\tab Natural language processing (NLP) is a discipline that aims to understand written or spoken words, similar to what humans do. NLP finds a place in various data-driven processes for a variety of subjects, including translating text from one language to another~\cite{lewis2020bart}, answering questions~\cite{wang2018multi}, summarization~\cite{zhang2024benchmarking}, digital assistants~\cite{king2023voice}. Besides these tasks, text classification, which assigns a document to a specific category, is one of the main areas in which NLP is employed. Categorizing customer requests, sorting emails or papers, understanding user ideas in social media, exploring trends, and topic labelling for documents are some of the application areas of text classification~\cite{rajendran2024local, shah2023review, salloum2022systematic, patton2020contextual, wang2023research}. In all these tasks, NLP builds a bridge between linguistic structure and computer-based analysis to cover the underlying conceptual content of large amounts of text data.

\tab NLP, or text processing, is a rapidly evolving area of computer science. The traditional text analysis methods depend on rule-based statistical modeling~\cite{chan2021natural}, such as Naïve Bayes, K-nearest, Decision Trees. They use hand-crafted feature engineering techniques to solve problems. Furthermore, classical algorithms are largely domain-dependent; it is important to have domain knowledge of the document beforehand to classify them. Therefore, they may not capture the intricate structural features and the context-dependent relationships between words and documents while learning textual representations~\cite{hemmatian2019survey, li2022survey}. For this reason, some of the early methods in text classification even require expert knowledge for labeling.

\tab Machine learning (ML) algorithms have provided a better platform for NLP and text-based classification problems. The key advantages of ML algorithms, such as automatic feature learning~\cite{bishop2006pattern} and capturing semantic and complex relations between words~\cite{jordan2015machine}, have become a remedy for overcoming difficulties in text classification and general NLP. Among the commonly used methods of ML-based algorithms, Bag of Words (BOW)~\cite{zhang2010understanding}, TF-IDF~\cite{aizawa2003information}, and n-grams~\cite{sidorov2014syntactic} were more prominent. Although successful, they also have some disadvantages.

\tab The success of machine learning algorithms depends significantly on their use of large amounts of data. Therefore, the increasing amount of data, which is a challenge in classical algorithms, is vitally important here. This also meets the requirement to represent the real world by sampling all possible data with sufficient diversity~\cite{sarker2021machine}. Therefore, one of the top priorities in machine learning algorithms is the quality and quantity of the data used.

\tab The classification is a process of questioning whether the text belongs to one of the predetermined categories. To perform such a classification task, knowledge of critical properties is necessary. These fundamental properties are often hidden in the text. It is often impossible to create a theoretical rule to determine which class the text will belong to~\cite{li2022survey} in advance. Moreover, the language is evolving and changing over time. Hence, the requirements related to classification are dynamic~\cite{he2018time}. Therefore, a dynamic learning algorithm is essential to understand the text's hidden features.

\tab The Grounded theory involves the application of inductive reasoning~\cite{birks2022grounded}, which is well suited to the document classification problem. Inductive reasoning implies that a model trained with enough sample data to learn the structure of language can be used to classify other documents without further training. The success of the model depends on the number of examples in the dataset. The more data collected, the closer the model will be to the real world. So, the model emerges from the data. A crucial aspect of successful data collection is that each category must be equally represented in the data set~\cite{coyne1997sampling}.

\tab The proposed study introduces a new method for text embedding. The merit of the proposed method is to extract the basic rules of the language from the data, which constitutes text documents. In this sense, it introduces a model based on qualitative research~\cite{merriam2015qualitative}. The proposed random walk-based embedding model is designed to extract semantic features of the sentences from the text-based material through inductive learning and creating a universal word graph. The word graph in which nodes are words, and the weighted links are connections between them is used to generate a transition probabilities-probability of the closeness of the word in sentences. In such a graph, some words take a central (hub) role due to their subject-based importance. The subject-related hub (key) words in the universal (covering all words of a language) weighted graph of words shape the graph topology, resulting in unique embedding vectors for each word emphasizing consecutive arrangement of words and several co-occurrences in text. Such a graph consists of the words (nodes) and semantic relations (links) among the words. Providing that the theoretical sampling is achieved, the resulting graph contains all possible words to represent the current language.

\tab The proposed embedding method is based on the Transition Probability Matrix (TPM) method~\cite{n2023tpm}. The TPM method calculates embedding vectors from the transition probabilities obtained employing random walks on the graph. In the proposed model, the graph is constructed using words, and edge weights are assigned based on the semantics of the training text. Hence, random walks are performed on the weighted graph with non-equal probabilities for choosing the neighboring nodes. For this reason, the new approach is called the Guided Transition Probability Matrix (GTPM) model. In this work, we explain the proposed algorithm and present a comparative study of its success in text classification. Our results show that the GTPM algorithm accurately captures the semantic meaning of the text, which is reflected in the successful classification results.

\tab The rest of the paper is organized as follows: the next section discusses the related work. Section 3 demonstrates how to construct a weighted word graph by analyzing word positions within documents. Comprehensive experiments are performed to verify the proposed model in Section 4. Section 5, is devoted to discussions; the paper concludes and discusses the future work.

\section{Related Work}
\label{RelatedWork}

\tab Unlike the numerical data, the text data contains a wealth of information. There are three types of information embedded in the text data: Information on the relations between the words, the semantics of the sentences, and the content of the text. Nevertheless, this rich content can not be used directly in machine learning models. Extracting the rich structure of text and transforming it into quantitative data remains one of the most difficult challenges for machine learning researchers. Reaching all this information requires a series of complex tasks, advanced techniques, and expert supervision. Two of the essential processes to convert text data into numerical representations are word embedding and feature extraction. The word embedding is the process of extracting relationships between words through sentence structure.

\tab 
A sentence can be seen as a walk among the words. The steps and the walk length are given by the sentence structure. The first deep learning methods for graph embedding have been started with DeepWalk~\cite{perozzi2014deepwalk}, node2vec~\cite{grover2016node2vec}, struct2vec~\cite{li2016discriminative}, and HARP~\cite{chen2018harp}. These methods use random walk-based embedding techniques and result in feature extraction.

\tab Deep learning models have significantly improved text classification performance by capturing meaningful word patterns. However, text data keeps hidden relations between words and documents. The above approaches are sometimes insufficient because they focus on local word relations and may struggle with long-term dependencies and global semantics in complex texts. To this end, the proposed method to deal with this situation is discussed in the next section.

\tab The text data contains more information than the information extracted using sentences. The sentence structure, a rule-based connectivity among words, implies that the best representation of the text data is a graph of words. Connectivities among the words and the closeness of words in different sentences create a graph structure.

\tab A recent publication has shown that the local topological structure around the nodes of a graph can be obtained in terms of transition probabilities between the neighboring sites~\cite{n2023tpm}. The transition probabilities obtained from anonymous walks starting from a particular node are used to construct a transition probability matrix used as the feature vector of the node in concern. The method has shown its success as a very efficient feature embedding method, and it is called the Transition Probability Matrix (TPM) method~\cite{n2023tpm}.

\tab Even though the text data and graph data are seemingly different, the inherent relationships between the words help to organize the graph structure of the words of a text. In this sense, the TPM method can be adopted for the investigations of text data. The generalized version of the TPM method will be presented in the following section.

\section{Method}
\label{Method}

\tab Given that the semantics of words are closely related to neighboring words, graph structures naturally provide an ideal framework to extract these relationships. Hence, selecting an effective strategy for text representation as graph data is vital. Successful feature extraction algorithms often rely on random walks within a document. These walks help uncover the underlying relationships between words, which allows for the representation of semantic information using a vector of numerical data. The celebrated "node2vec" algorithm~\cite{grover2016node2vec} is a notable example of leveraging this concept to derive feature vectors from the text.

\tab A language is complete with words, syntax, and semantics. As social, economic, and technological changes appear in society, the language gains new words and ways of expression. In this sense, each field of interest develops its own jargon and characteristic words, similar to keywords. A new social phenomenon enriches the language and, at the same time, heavily influences word positioning relative to these key terms, creating a signature of the field, which becomes fingerprints for each document. Also, the new word sequences and their appearance frequencies set new structural rules of the language. Even though the same language is used by every individual regardless of their profession, the choice of words and sentence structures reveals the hidden context of the text.

\tab The question appears that despite the same language being used in all aspects of life, how can one extract deeper information by using the characteristic words that play hub roles in connecting more commonly used words? Colloquial language is a common ground for communicating, while technical or subject-specific words are used to express the nuances of the subject matter. Therefore, the primary aim of the proposed study is to include and emphasize the uses and connotations of words in different fields in the feature extraction stage of NLP.

\tab A two-step feature extraction algorithm is proposed in the present work. The steps are:

\begin{enumerate}
\item Creation of a universal weighted word graph.     
\item Embedding of any given document using the universal word graph.
\end{enumerate}

\subsection{Universal weighted graph of word}
\tab To exemplify the construction of the universal weighted graph of words, assume having a collection of sentences from a collection of documents that belong to different classes (for example, sports, news, science, etc.). $w_i$ and $d_i$ represent words in sentences and documents, respectively. Words represent the nodes of the graph without any connection. As the documents are processed, connectivities among the words start to emerge according to the wordings of the sentences; a connected graph emerges as a consequence of the word structure of the documents.

\begin{figure}[h]
   \centering
  \includegraphics[width=0.99\textwidth]{"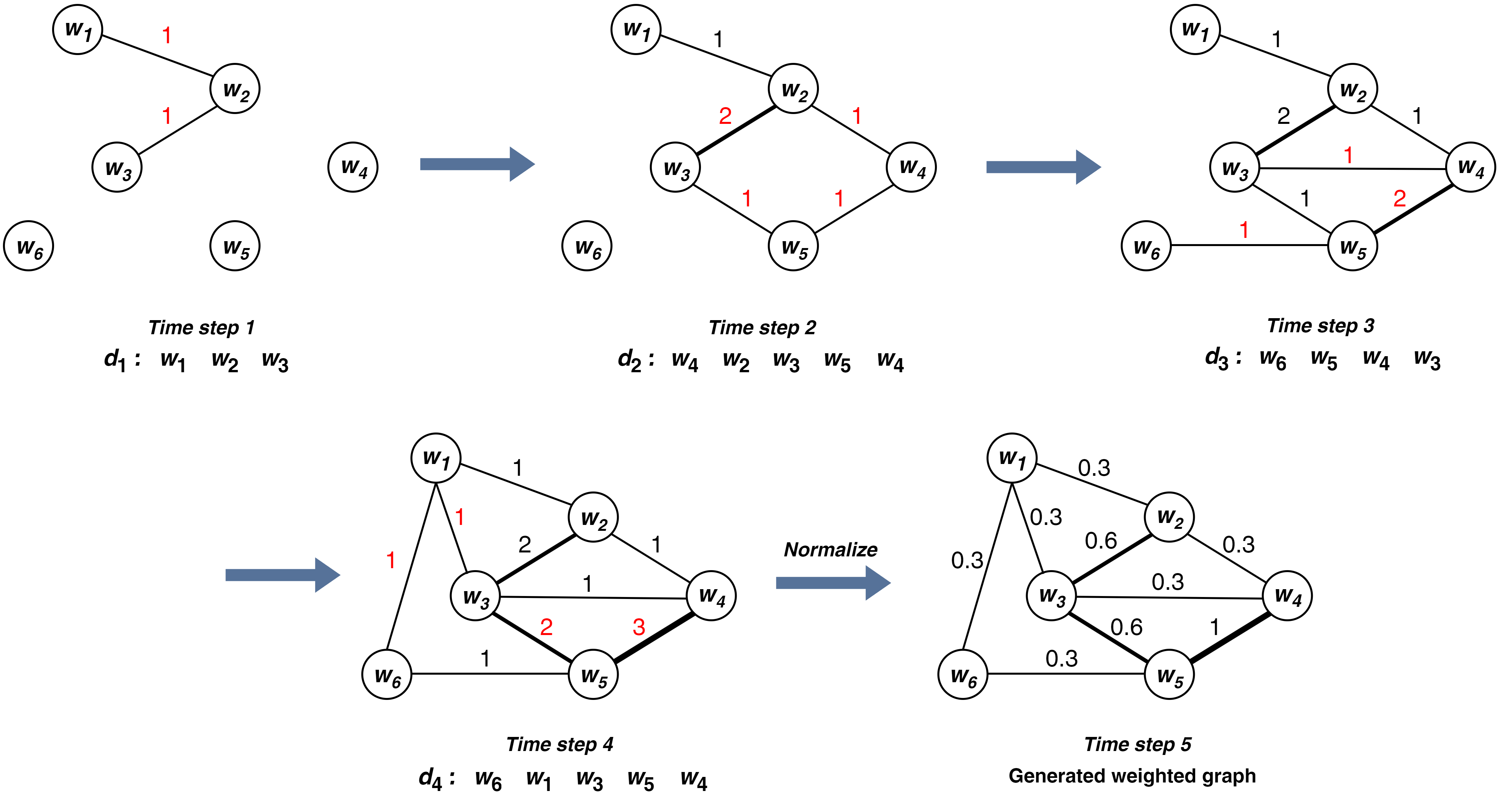"}
\caption{Generation of word graph for four iterations. At time step 1, the first document is introduced, and a graph with three words is generated. At time step 2, a second document is introduced, which adds two new nodes to $w_4$ and $w_5$, increases the edge value between $w_2$ and $w_3$, and so on. In the last time step, all edges are normalized with the maximum edge value. }
   \label{WordsToGraph}
  \end{figure}
  
\tab Figure~\ref{WordsToGraph} illustrates the process using four small documents from four categories. The wording of four documents creates a six-word language. Among these six words, there is no connection before processing the documents; therefore, the connectivity matrix, $A_i{}_j$ is defined as, $a_i{}_j=0$, $i = 1,2,\dots 6$, $j = 1,2,\dots 6$. 

\begin{table}[H]
\centering
\begin{tabular}{clcccclc}
\cline{2-2} \cline{7-7}
\multicolumn{1}{c|}{} & & $a_{11}$ & $a_{12}$ & \dots  & $a_{1j}$ & \multicolumn{1}{l|}{} \\ [1.5ex]
\multicolumn{1}{c|}{} & & $a_{21}$ & $a_{22}$ & \dots  & $a_{2j}$ & \multicolumn{1}{l|}{} \\ [1.5ex]
\multicolumn{1}{c|}{$A_{ij} =$} & & \dots     & \dots     & \dots  & \dots     & \multicolumn{1}{l|}{} \\ [1.5ex]
\multicolumn{1}{c|}{} & & $a_{i1}$ & $a_{i2}$ & \dots  & $a_{ij}$ & \multicolumn{1}{l|}{} \\ [1.5ex]
\cline{2-2} \cline{7-7} 
\end{tabular}
\end{table}

\tab Whenever two words , $i$ and $j$, are consecutive in the text, corresponding element of matrix, $a_i{}_j$, is increased by one. 

$$
a_{ij} =
\begin{cases} 
a_{ij} = a_{ij} + 1 & \text{\textit{if existing}, increase edge value} \\
a_{ij} = 1 & \text{\textit{if not existing}, add to the graph}
\end{cases}
$$

\tab At the end, $a_{ij}$ represents the number of co-occurrences of word $i$ with its consecutive neighbor $j$ in each sentence.

\tab Assume that we have the following documents which belong to different classes (Table~\ref{Documents}). Each document is introduced to the graph at each iteration.

\begin{table}[h]
\centering
\caption{Sample documents with corresponding IDs and types.}
\label{Documents}
\begin{tabular*}{\columnwidth}{
  @{\extracolsep{\fill}\hspace{\tabcolsep}}
  ccc
  @{\hspace{\tabcolsep}}
}
\hline
Document                                                                                &   Document ID     & Document Type     \\ \hline
$w_1 \hspace{0.3em} w_2 \hspace{0.3em} w_3$                                             &   document1       & $(d_1)$           \\ 
$w_4 \hspace{0.3em} w_2 \hspace{0.3em} w_3 \hspace{0.3em} w_5 \hspace{0.3em} w_4$       &   document2       & $(d_2)$           \\ 
$w_6 \hspace{0.3em} w_5 \hspace{0.3em} w_4 \hspace{0.3em} w_3$                          &   document3       & $(d_3)$           \\
$w_6 \hspace{0.3em} w_1 \hspace{0.3em} w_3 \hspace{0.3em} w_5 \hspace{0.3em} w_4$       &   document4       & $(d_4)$           \\ \hline

\end{tabular*}%
\end{table}

\tab As a characteristic of languages, some words appear more often. The frequently appearing words are divided into two groups. The first group is punctuation marks, suffixes, conjugation markers, and conjunctions such as, and, or. In preparation of the text data for any NLP model, these must be extracted from sentences. Another group of frequently occurring words are the "essential" words (jargon) of the subject, which play a decisive role in clarifying the sentence and giving it its distinct meaning. The main difference between the present approach and the earlier ones is the method of treating the keywords of the subjects. Instead of ignoring the repetition of the characteristic words of the topic, their contribution is strengthened: The strength of the connection between the characteristic words of the topic and the neighboring word increases each time two words are consecutive. The normalization process, including all edges, gives a weighted graph with the topic-specific characteristic words as hub words. After this stage, a random walk obeying the probabilities based on the link value (Figure~\ref{WordsToGraph}) is used to extract features of the nodes. The extracted node features inherently exhibit the semantics of the topic of interest.

\tab The steps of the proposed algorithm are:

\begin{enumerate}
\item Determining the universal dictionary. Extract all possible words using all available documents.      
\item Determining the connectivity structure. Elements of the adjacency matrix are the links between the words. Using all documents, obtain the connections between the nodes. The steps of creating the adjacency (connectivity) matrix:
\begin{itemize}
    \item Each time a new connection is observed, the connection is established; the relevant element of the adjacency matrix is set to 1.
    \item Each time an existing connection occurs, the link is strengthened; the relevant element of the adjacency matrix is increased by 1.
\end{itemize}

\end{enumerate}

\tab At this point, a "universal graph" containing all connectivity information within the predetermined language limits is available for extracting further data. This process is dynamic: Introducing new documents into the set of documents extends or enhances the connections of the graph. The above-described graph generation algorithm depends on the growth mechanism at each iteration, which is the main principle of the preferential attachment~\cite{barabasi2013network} growth mechanism of the graph generation. The observed power law behavior is also consistent with Zipf's law~\cite{powers1998applications}, which mathematically formulates the frequencies of words in the language. The law indicates a scale-free degree distribution for words and their rank.

\tab The most important aspect of this growth process is that, at each step, some nodes gain a more dominant (hub) role. Communities will appear on the graph if the documents are collected on different subjects. The communities are densely connected regions of subject-related words, and the topic-specific keywords are the hub nodes.

\begin{figure}[h]
   \centering
  \includegraphics[width=0.8\textwidth]{"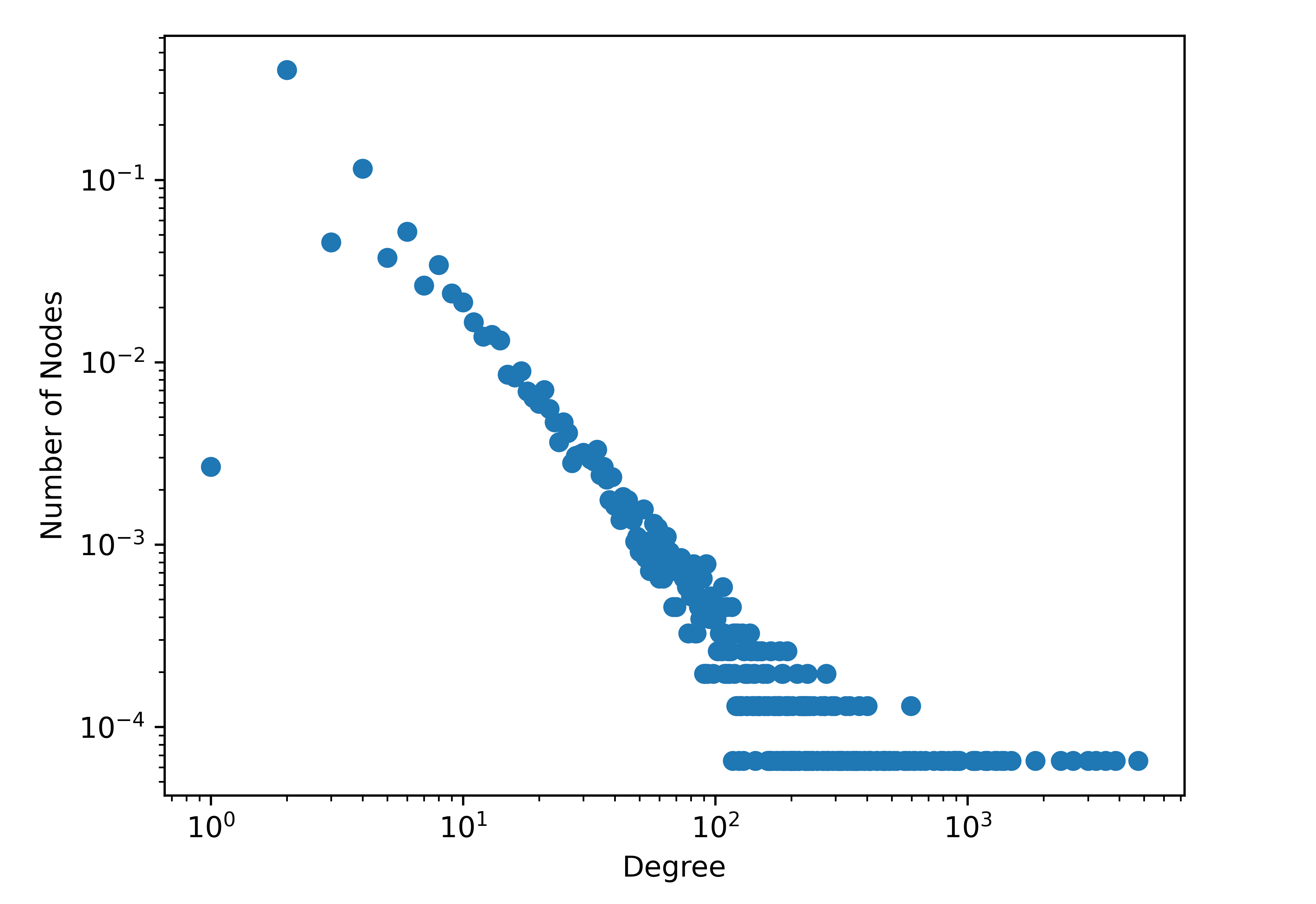"}
\caption{Degree distribution of words in the generated word-graph using the Reuters dataset. }
   \label{DegreeDistribution}
  \end{figure}

\tab To test the created network structure, we have used Reuters (Keras) dataset~\cite{chollet2021deep} which contains 11,228 documents written in English, having 10,054 words (words which appear at least 5 times). The documents are on 46 different subjects. Figure~\ref{DegreeDistribution}  shows the degree distribution of the universal word graph obtained by using Reuters dataset. Each subject has some hub words that connect a sub-graph of words, which exhibits the power law behavior consistent with the scale-free networks in complex-network theory.

\tab Power law behavior for language was introduced to the literature by the linguist George Kingsley Zipf in 1935, which mathematically formulates the frequencies of words in the language, called Zipf's law ~\cite{powers1998applications}. This law has a scale-free degree distribution for words and their rank. The degree distribution of the created graph is compatible with Zipf's law. This proves that enough samples have been made to understand the language's structure and represent the diversity of words in the language. Also supports the compatibility of the graph creation strategy.

\subsubsection{Weighted random walk}
\tab The feature extraction process starts with an adjacency matrix representing the universal word graph. Random walks on the weighted word graph exhibit the relative importance of the connections between the words. Hence, a random walk starting from a node contributes to the creation of the node feature vector of the node that respects the relative importance of the word within the document.

\tab In an unweighted graph, the probability of a walker $i$ visiting one of the nearest neighbors $j$ has an equal probability. Equal probability omits the relative importance of the connection; the weights contain the relative probability information. The weights on the edges, the elements of the adjacency matrix, are normalized; hence, the elements of the adjacency matrix provide the probabilities of random walks.

\tab In this work, the probabilities of the random walk is given as,

\begin{equation}
p_{i, j}=\frac{e_{i j}}{\sum_{j \in N(i)} e_{i j}}
\end{equation}

\tab Here $e_{ij}$ represents the edge weight between nodes $i$ and $j$ , which is $a_{ij}$, and denominator term is, $\sum_{j \in N(i)} e_{i j}$, the normalization.

\bigskip

\tab The probability of visiting a neighbor on a random walk depends on that neighbor's edge weight. The stronger the link, the more likely it is to be selected. So, the number of co-occurrences biases this selection process to choose the more closely related neighbor. For this reason, embedding vectors are expected to contain information not only about the relationship between words but also about the neighborhood information resulting from the topological structure.

\subsection{Embedding of any given document using the universal word graph}

\tab The random walks are further processed to obtain embedding vectors. In the proposed model, the embedding technique is akin to the one already introduced in a previous work~\cite{n2023tpm}. The section~\ref{RelatedWork} contains a summary for completeness.

\tab The random walks starting from the node in concern are the best starting point for collecting all possible information related to connectivities and the local topology structure of graphs. The feature vector of a node, obtained using the information collected by the random walks, contains information about the connectivity with its neighbors in the periphery of the node.

\tab Each random walk starting from the same node collects different characteristics of the local neighborhood of this node; hence, relations are mapped to the numerical representation as the features of the nodes to obtain embedding vectors. This approach is common for all random walk-based feature extraction processes. A shortcoming of this approach is that the walks convey information specific to particular neighborhoods; feature vectors contain local identities. Despite the feature vectors containing local information, the characteristics of a scale-free graph are universal. This property of random walks prevents the generalization of obtained topological information.

\tab A recent method of embedding based on anonymous random walks~\cite{n2023tpm} has shown that one can map the local connectivity relations as a transition probability matrix that maps the topological structure of the neighborhood without any reference to the identities of the nodes in concern~\cite{ivanov2018anonymous}. The information embedded into anonymous walks contains only the relationships among neighboring nodes.

\tab Closely following the approach presented in~\cite{n2023tpm}, the embedding method consists of the following steps:

\begin{enumerate}
    \item The local topological structure of the neighborhood of a node is searched by performing random walks of predetermined length.
    \item Each random walk is transformed into anonymous walks~\cite{ivanov2018anonymous}.
    \item The anonymous walks are converted into a transition probability matrix.
\end{enumerate}

\tab The main difference between the original work, which discussed the Transition Probability Matrix, and the present work is that in the present work, random walks are realized on a weighted graph, whose weights are the probabilities of choosing any neighbor at each step of the walk.

\begin{figure}[h]
   \centering
  \includegraphics[width=0.99\textwidth]{"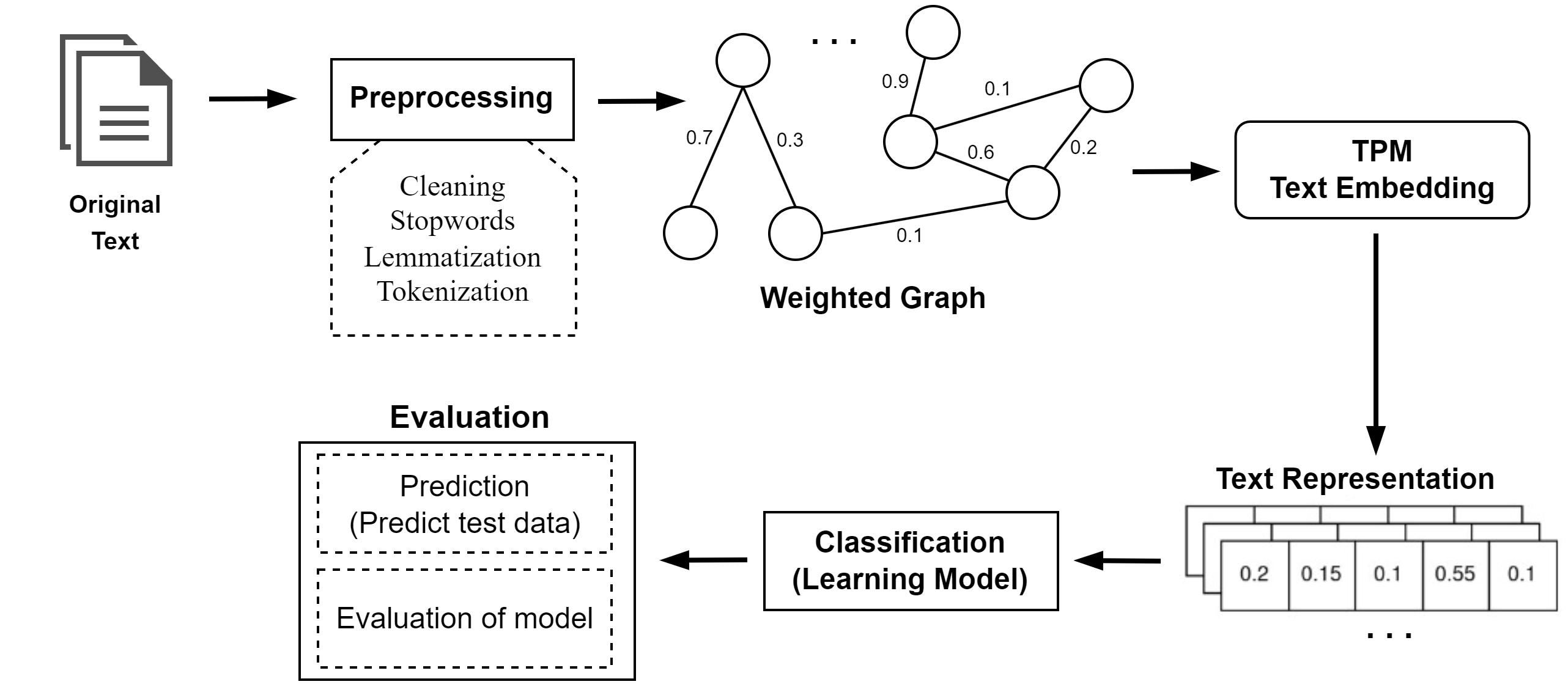"}
\caption{Representation of the study. }
   \label{Framework}
  \end{figure}

\tab Figure~\ref{Framework} shows the process of text classification using the proposed embedding algorithm.

\tab The resulting probability matrix represents the topology of the node’s relationships with its neighbors, essentially serving as its signature. This approach offers two significant benefits: firstly, it aggregates information encompassing all features of the network’s relationships within the node, and secondly, it elucidates the similarity between the node’s features and those of nodes sharing similar topological relationships.

\tab In this process, the number of walks and the length of the walks enable the aggregation of local and global information in the graph. The graph creation strategy ensures the link between the topological information, the rules of the language, and the relationships between words. Increasing the number of walks or walk lengths enables collecting more accurate information on the neighborhood. Therefore, high-quality embedding vectors can be obtained by adjusting these two parameters.

\section{Results}

\tab This section is devoted to the text classification performance of the proposed embedding algorithm. Six datasets, SST-2, MR, CoLA, Ohsumed, Reuters (Keras), and 20NG, are used to test the classification performance of the embedding algorithms. The first three datasets are used for binary classification, while the rest are multi-class data sets used for the more challenging task of multi-class classification. The classification performances of the proposed and commonly accepted feature embedding algorithms have been subject to comparison using the same datasets.

\tab The pre-processing of datasets, including cleaning texts and tokenizing words, is realized using the NLTK library\footnote{\url{https://www.nltk.org/} [Accessed date: 28 December 2023]}. The node feature vectors, the results of the embedding processes, are used for text classification. The proposed embedding algorithm is used in conjunction with a multi-layer neural network. The details of the neural network will be described in the following sub-section.

\tab Eight well-known baseline models and successful embedding classification algorithms are employed for the comparisons. For all eight embedding algorithms, the default parameter settings are used as described in their original papers or implementations. We employed 300-dimensional BERT word embedding for models utilizing pre-trained word embedding. The experimental findings presented in this study are the average of five runs under the same preprocessing.

\tab For a comprehensive evaluation of the models, we utilized the two well-known metrics, Micro-F1 and Macro-F1, to evaluate the performance of all the methods.

\begin{itemize}
  \item \textbf{Micro-F1} is a metric considering the overall precision and recall of all the labels~\cite{schutze2008introduction}.
\end{itemize}

\begin{equation}
\text { Micro-F1 }=\frac{2 P_s \times R_s}{P+R}
\end{equation}

where 

\begin{equation}
\text { Precision }(P)=\frac{\sum_{s \in \mathcal{S}} T P_s}{\sum_{s \in S} T P_s+F P_s}, \quad \text { Recall }(R)=\frac{\sum_{s \in S} T P_s}{\sum_{s \in \mathcal{S}} T P_s+F N_s}
\end{equation}

\begin{itemize}
  \item \textbf{Macro-F1} metric calculates the F1 score for each class label separately and then takes the average of all the F1 scores~\cite{schutze2008introduction}. In contrast to Micro-F1, which assigns equal weight to every instance, Macro-F1 assigns equal weight to all labels during the averaging process.
\end{itemize}

\begin{equation}
\text { Macro-F1 }=\frac{1}{\mathcal{S}} \sum_{s \in \mathcal{S}} \frac{2 P_s \times R_S}{P_S+R_S}
\end{equation}

where

\begin{equation}
P_s=\frac{T P_s}{T P_s+F P_s}, \quad R_s=\frac{T P_s}{T P_s+F N_s}
\end{equation}

\bigskip
\bigskip
where $S$ represents the number of classes in a multi-class classification problem. $P_s$ and $R_s$ represent precision of a specific class $s$ and recall of a specific class $s$. $TP_s$, $TN_s$, $FP_s$, and $FN_s$ represent the number of true positives, true negatives, false positives and false negatives, respectively, for category $s$ within the set of classes $S$.

\subsection{The Datasets}
\tab Table~\ref{StatisticsOfDatasets} shows the information, which includes the number of documents, the number of classes, the average word length, and the related field of the datasets.

\begin{table}[]
\centering
\caption{Statistics of datasets.}
\label{StatisticsOfDatasets}
\resizebox{\textwidth}{!}{%
\begin{tabular}{ccccccc}
\hline
\textbf{Dataset} & \textbf{$\#$ Docs} & \textbf{$\#$ Training} & \textbf{$\#$ Test} & \textbf{$\#$ Classes} & \textbf{Avg. Length} & \textbf{Explanation} \\ \hline
\textbf{SST-2}   & 8,741            & 6,920                & 1,821            & 2                   & 19.3                 & Movie reviews        \\
\textbf{MR}      & 10,662           & 7,108                & 3,554            & 2                   & 21                   & Movie reviews        \\
\textbf{CoLA}    & 10,657           & 9,594                & 1,063            & 2                   & 7.7                  & Grammar dataset      \\
\textbf{Ohsumed} & 7,400            & 3,357                & 4,043            & 23                  & 79.49                & Medical abstracts    \\
\textbf{Reuters} & 11,228           & 8,982                & 2,246            & 46                  & 60.28                & Newswires            \\
\textbf{20NG}    & 18,846           & 11,314               & 7,532            & 20                  & 221.26               & Newswires            \\ \hline
\end{tabular}%
}
\end{table}

\tab The above-described datasets are used to create the embedding vectors using nine different algorithms, including the proposed algorithm; details are described in the section~\ref{Method}. Before applying the embedding algorithms, all datasets are subject to the data cleaning process. The sentences are stripped from the artifacts until the base of the word remains using the NLTK library. 

\tab To obtain node feature vectors, the proposed algorithm (GTPM) applies the following procedure to each dataset: After the cleaning, the distinct words become nodes of the graph. Connections and the connection weights are calculated using the information embedded in the original text. Random walks were performed on the weighted graph to calculate the transition probability matrix (TPM)~\cite{n2023tpm}. The obtained transition probability matrix is converted to distinct feature vectors for each node, representing the local topology and connectivity of the starting node.

\subsection{Baseline Models}

\begin{itemize}
  \item \textbf{TF-IDF + LR:} In the process of feature extraction, Term Frequency-Inverse Document Frequency (TF-IDF) is employed to transform the textual documents into feature representations. These TF-IDF-based features are subsequently employed as inputs for the Logistic Regression (LR) classifier.
\end{itemize}

\begin{itemize}
  \item \textbf{CNN-pretrain:} The Convolutional Neural Network (CNN), as introduced by~\cite{kim-2014-convolutional}, conducts convolution and max-pooling operations on pre-trained word embeddings (BERT) to perform text classification.
\end{itemize}

\begin{itemize}
  \item \textbf{Bi-LSTM:} Bidirectional Long Short-Term Memory (Bi-LSTM) is a type of recurrent neural network architecture designed to address the vanishing gradient problem common in standard RNNs. Bi-LSTMs process sequence data in both forward and backward directions, enabling the capture of long-term dependencies. For this study, a Bi-LSTM model was implemented utilizing pre-trained Bidirectional Encoder Representations from Transformers (BERT) word embeddings as input features. The final hidden state of the Bi-LSTM was used for text classification. 
\end{itemize}

\begin{itemize}
  \item \textbf{SWEM:} A Simple Word Embedding Model (SWEM)~\cite{shen2018baseline} features basic pooling techniques applied to word embeddings. We utilized the SWEM-average approach in our experimental setup, incorporating pre-trained embeddings from BERT as features. We classified text using an architecture consisting of layers with configurations (128-256-512-C), where C represents the number of distinct classes.
\end{itemize}

\begin{itemize}
  \item \textbf{TextGCN:} Text Graph Convolutional Network (TextGCN)~\cite{yao2019graph} models the text corpus as a single interconnected graph with words and documents as nodes and document-word and word-word co-occurrence as edges, then utilizes graph convolutional networks to perform node classification on this graph representation for text classification.
\end{itemize}

\begin{itemize}
  \item \textbf{TextING:} Text classification method for INductive word representations via Graph neural networks (TextING) constructs separate graphs for each document and utilizes a gated graph neural network to learn text-level word interactions, aiming to improve its ability to preserve both local and global contextual information in an inductive learning fashion~\cite{zhang2020every}.
\end{itemize}

\begin{itemize}
  \item \textbf{BERT:} The BERT model architecture utilizes bidirectional Transformer encoders to represent each token in a textual input sequence while incorporating contextual information from the full surrounding input sequence. This enables BERT to generate contextualized token representations bidirectionally from the input transformers~\cite{devlin2019bert}. 
\end{itemize}

\begin{itemize}
  \item \textbf{VGCN-BERT:} Vocabulary Graph Convolutional Network-BERT (VGCN-BERT) model uses a simple graph convolutional network to extract features from a vocabulary graph to enhance word embeddings, which are input to the self-attention encoder in BERT for classification~\cite{lu2020vgcn}.
\end{itemize}

\subsection{Two-dimensional mapping of classes}
\begin{figure}[h]
  \centering
\begin{subfigure}{.32\textwidth}
  \centering
  \includegraphics[width=1\linewidth]{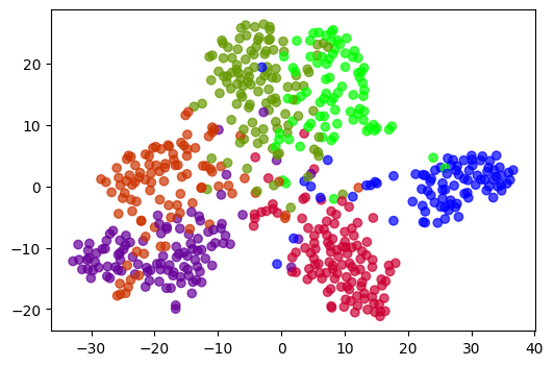}  
  \caption{TextGCN}
  \label{Vis_DW}
\end{subfigure}
\begin{subfigure}{.32\textwidth}
  \centering
  \includegraphics[width=1\linewidth]{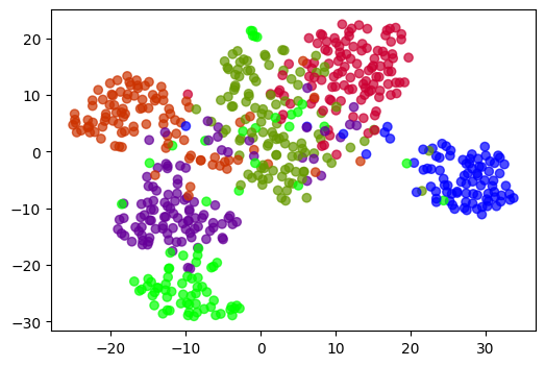}  
  \caption{TextING}
  \label{Vis_GCN}
\end{subfigure}
\begin{subfigure}{.32\textwidth}
  \centering
  \includegraphics[width=1\linewidth]{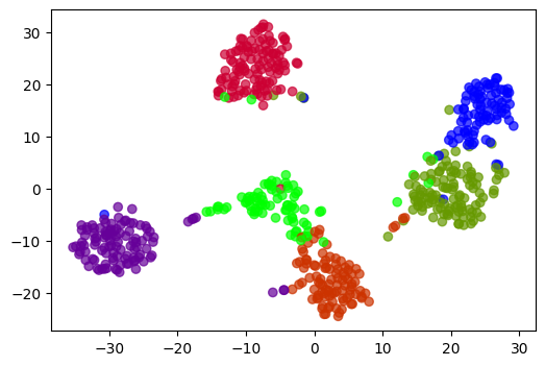}  
  \caption{GTPM}
  \label{Vis_TPM}
\end{subfigure}
\caption{Visualization of 2D representations for Reuters dataset.}
\label{Visualization}
\end{figure}

\tab The visual inspection is one of the best tests for classification algorithms. Mapping multi-dimensional vectors to two- or three-dimensional vectors is an essential process for visualization. Dimensional reduction is a common practice to reduce a multi-dimensional array to low-dimensional, graphically representable arrays. Moreover, reducing the dimensions of the feature vectors eliminates the artifact of classification algorithms and emphasizes the potential of the embedding algorithm. For the Reuters dataset, feature vectors are obtained using three different (graph-based) embedding methods (TextGCN, TextING, proposed model (GTPM)). The obtained feature vectors are reduced to 2 dimensions using the TSNE algorithm~\cite{van2008visualizing}. Figure~\ref{Visualization} shows that each embedding vector, belonging to different classes, is grouped by itself and separated when compared with the other methods.

\subsection{Text classification Process}
\tab The feature vectors obtained using the proposed embedding algorithm are used as input for a multi-layer neural network for text classification. Keras API of TensorFlow is employed for the construction of sequential five layers [64, 128, 256, 512, output Layer], neural structure with ReLU activation function in each hidden layer, and either sigmoid or softmax activation function for the output layer, respective to the requirements of the particular classification task. Adam optimizer~\cite{kingma2014adam}, with the learning rate and dropout rates chosen from the sets [0.1, 0.001, 0.0001, 0.02, 0.002, 0.003], [0.1, 0.2, 0.5] respectively. Additionally, the early stopping strategy controlled the number of epochs. The data is divided into training and test sets for the text classification tasks. In each
case, $10\%$ of the training data was randomly chosen as the validation set.

\subsection{Parameter Selection and Robustness}
\tab The dimensional reduction results (Figure~\ref{Visualization}) show that GTPM can discriminate different classes comparatively better than other embedding algorithms. Hence, this observation increases expectations for the Guided Transition Probability Matrix (GTPM) approach, providing an excellent embedding basis for all text-processing purposes. All random walk-based embedding approaches carry an inherent dependence on two parameters. These parameters are the number of statistically independent walks from each node and the walk's depth (number of steps). The fine-tuning of these two parameters mainly depends on the network size. Another main concern of all embedding algorithms is the size of the training data. What is the relation between the size of the training set and the success rate? In this subsection, the number of walks and walk depth depend on the success of the GTPM embedding method and the robustness, which is the smallest training data set sufficient for comparable successful results. First, an embedding algorithm based on random walks improves as the number of walks increases. Is there a threshold at which increasing the number of walks contributes significantly less than previous numbers? Also, the depth of the walks is an influential factor. Any dependence on the depth of the walk and the nature of the text. Finally, robustness means how much training is enough for a good performance of the model. The following two subsections present parameter selection considerations regarding the robustness of model success and its dependence on training material.

\subsubsection{Optimal Parameter Selection}
\begin{figure}[ht]
  \centering
\begin{subfigure}{.45\textwidth}
  \centering
  \includegraphics[width=1\textwidth]{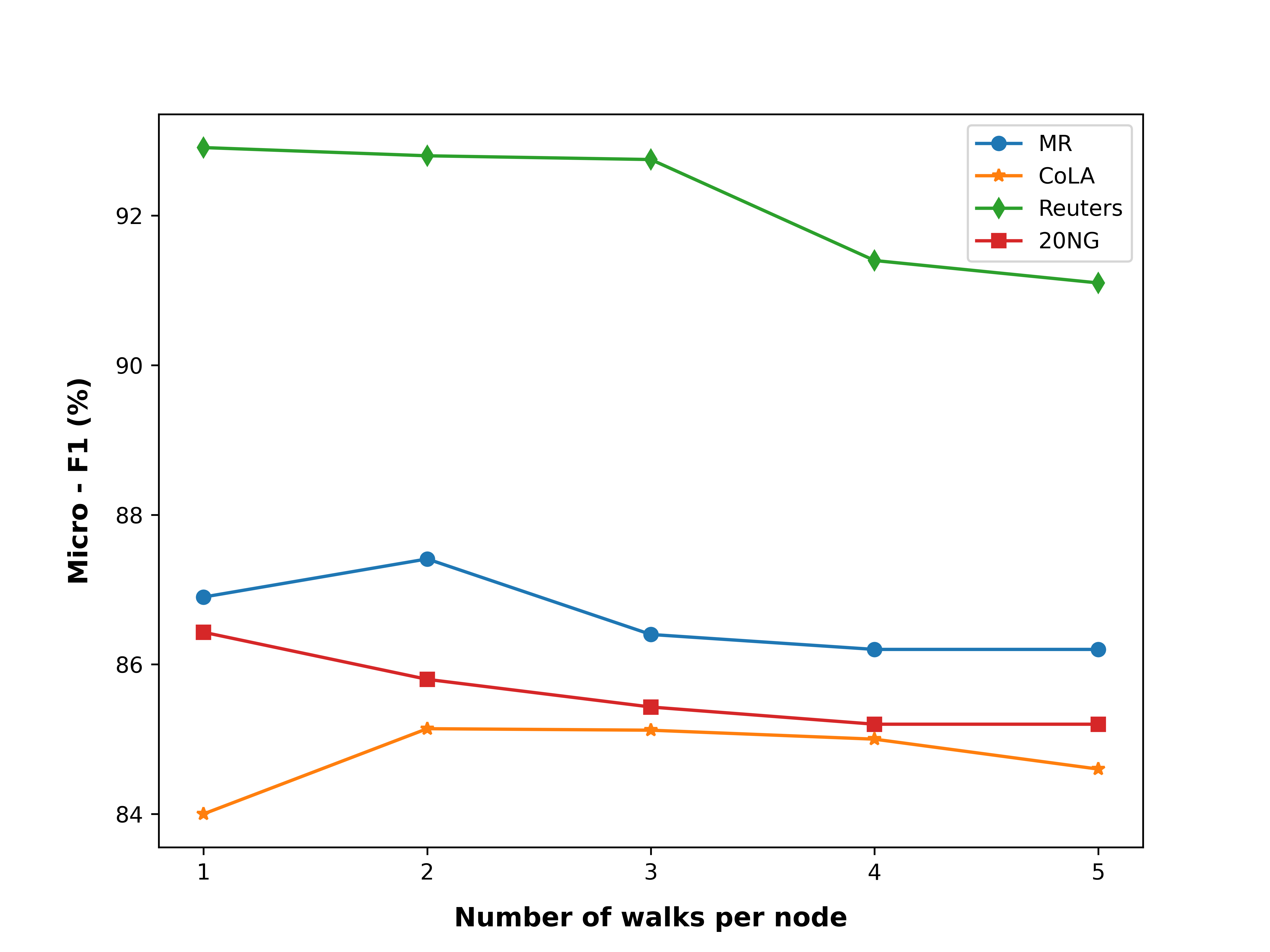}  
  \caption{}
  \label{PS_NW}
\end{subfigure}
\begin{subfigure}{.45\textwidth}
  \centering
  \includegraphics[width=1\textwidth]{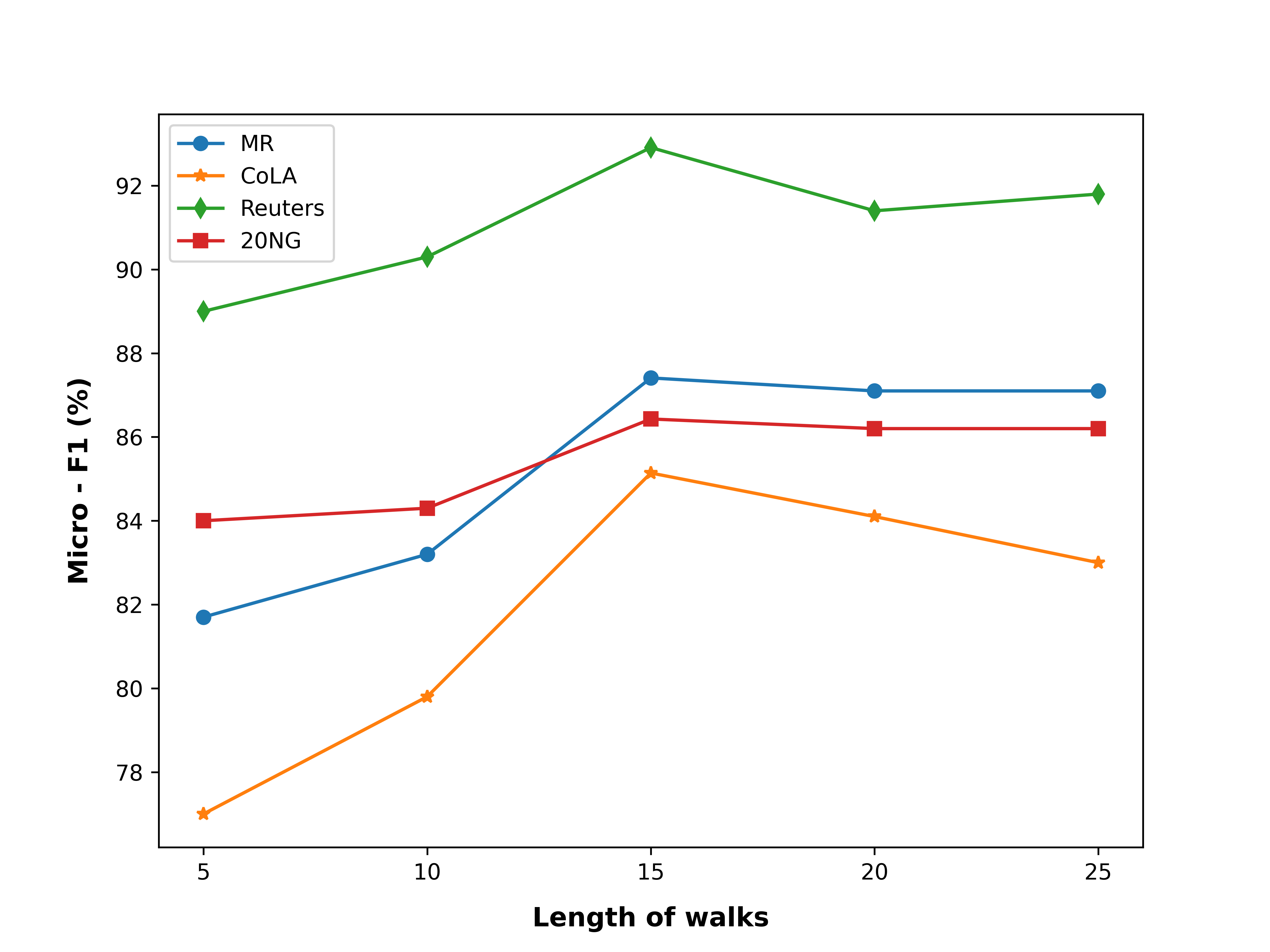}  
  \caption{}
  \label{PS_WL}
\end{subfigure}

\caption{Test accuracy (Micro-F1 $\%$) with different walk lengths ($a$) and number of walks per node ($b$).}
\label{Parameter_Sensitivity}
\end{figure}

\tab The proposed embedding algorithm, GTPM, includes two essential hyperparameters, $n$ and $m$, the number of walks per node and the length of walks. To find optimized values of these two parameters, walk lengths and the number of walks per node have been varied to observe the parameter dependence of performance. Figures~\ref{PS_NW} and~\ref{PS_WL} show the test results.

\tab The observed behavior is that the model's effectiveness improves with the increasing walk length up to a point, and it decreases after a peak. The model achieves optimal performance when the walk length is around 15 steps. The optimal number of random walks started from each node (word in a document) depends on the dataset’s average sentence length. For datasets with longer average sentence lengths, such as Reuters and 20NG, initiating a single random walk from each node yielded the best performance. Conversely, datasets with shorter average sentence lengths, such as MR and CoLA, require more than one random walk per word for better results. In the comparative studies of the success of the embedding, the length of walks is set to $m = 15$ for the GTPM. Similarly, the performances of all other embedding algorithms used for comparisons were measured using the optimal parameter values obtained from the literature.

\subsubsection{Robustness}
\tab The performance of an embedding algorithm under the limited training set is an important but, at the same time, not much-discussed measure. The robustness of the proposed model is tested against the above-mentioned embedding algorithms. The experiments utilized the Ohsumed and 20NG datasets with a constrained number of training data. Specifically, only $10\%$ of the entire training set is used while maintaining the original test set. As a result of limiting the training set, a significant portion of the words in the test set were not encountered during the training process. The outcome is illustrated in Table~\ref{Inductive}. In the comparative test, two graph-based baseline models (TextGCN and TextING) are also trained using the same training set. The success of the models is measured in terms of Micro and Macro F1 scores. The comparative results obtained using both datasets demonstrate a notable superiority of the GTPM method.

\begin{table}[!htbp]
\centering
\caption{Micro-F1 and Macro-F1 scores for the multi-class text classification datasets in inductive setting. Performance reductions from Table~\ref{MultiClassification} are denoted in parentheses.}
\label{Inductive}
\resizebox{\textwidth}{!}{%
\begin{tabular}{ccccc}
\hline
\multirow{2}{*}{\textbf{}}                                    & \multicolumn{2}{c}{\textbf{Ohsumed (10\%)}}                       & \multicolumn{2}{c}{\textbf{20NG (10\%)}}                          \\ \cline{2-5} 
                                                              & Micro F1                        & Macro F1                        & Micro F1                        & Macro F1                        \\ \hline
\textbf{TextGCN}                                              & \multicolumn{1}{l}{54.31 (-14)} & \multicolumn{1}{l}{44.07 (-17)} & \multicolumn{1}{l}{70.89 (-15)} & \multicolumn{1}{l}{63.39 (-16)} \\
\textbf{TextING}                                              & 58.63 (-13)                     & 54.11 (-15)                     & 63.52 (-19)                     & 63.27 (-19)                     \\
\textbf{GTPM}                                  & 70.26 (-6)                      & 60.97 (-10)                     & 82.67 (-4)                      & 80.21 (-6)                      \\ \hline
\multicolumn{1}{l}{\textbf{\# Samples/words in Training set}} & \multicolumn{2}{c}{335/5,409}                                     & \multicolumn{2}{c}{1,131/7,568}                                   \\
\textbf{\# Samples in Test set}                               & \multicolumn{2}{c}{4,043}                                         & \multicolumn{2}{c}{7,532}                                         \\ \hline
\end{tabular}%
}
\end{table}

\begin{figure}[H]
  \centering
\begin{subfigure}{.45\textwidth}
  \centering
  \includegraphics[width=1\textwidth]{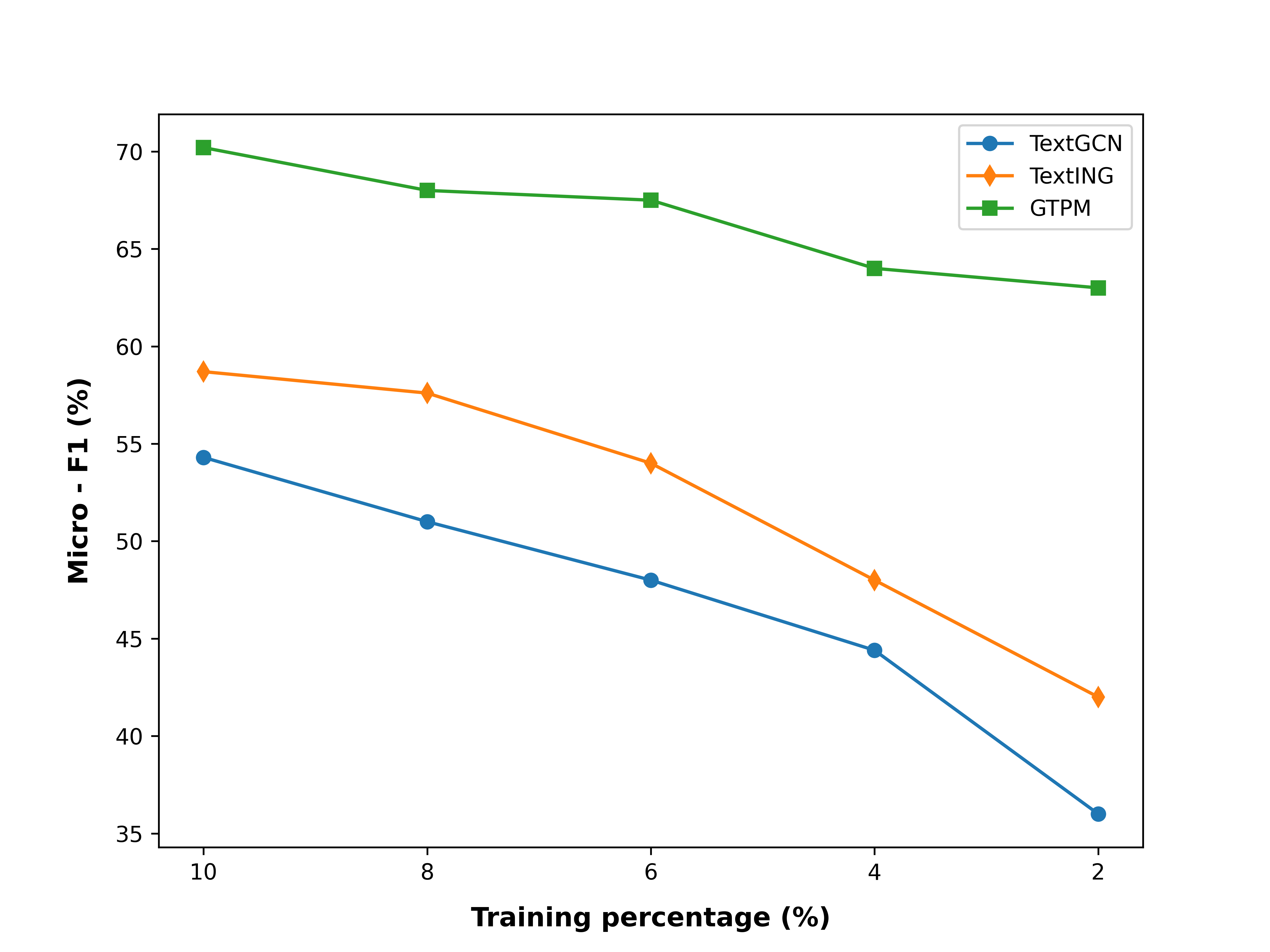}  
  \caption{}
  \label{Inductive_Ohsumed}
\end{subfigure}
\begin{subfigure}{.45\textwidth}
  \centering
  \includegraphics[width=1\textwidth]{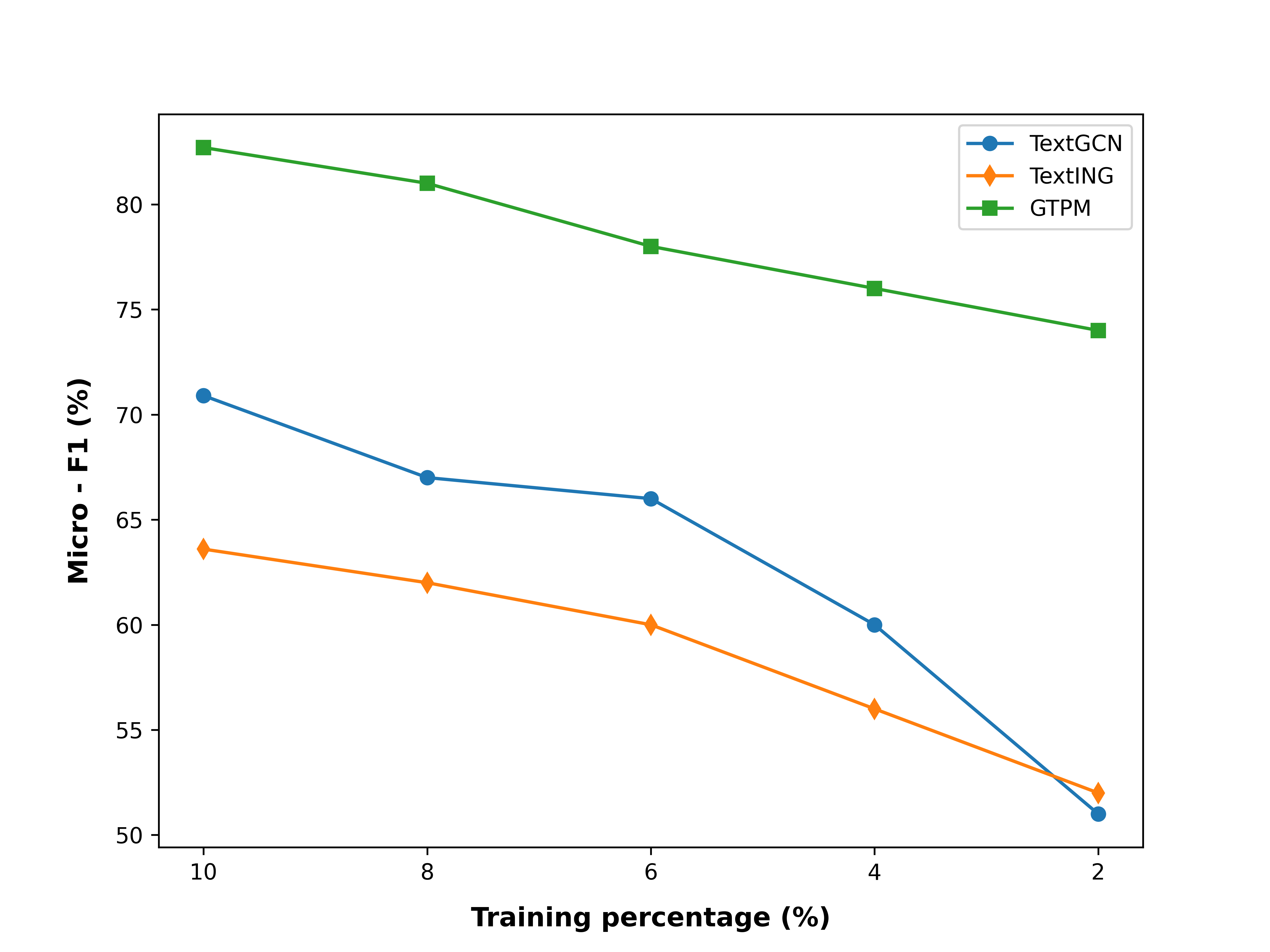}  
  \caption{}
  \label{Inductive_20NG}
\end{subfigure}

\caption{Test performance (Micro F1) with different train size on Ohsumed and 20NG datasets.}
\label{InductiveFigures}
\end{figure}

\tab Figure~\ref{InductiveFigures} illustrates the comparison of TextGCN, TextING, and GTPM methods on different proportions of labeled training data (ranging from $10\%$ to $2\%$) of the Ohsumed and 20NG datasets while maintaining the original test set size. The proposed method demonstrates consistent performance as the number of unseen words increases. Across both datasets, a decline in performance for the GTPM method is limited to approximately $8\%$, whereas other methods exhibit a more pronounced decrease ranging from $15\%$ to $20\%$. As a result, the GTPM method seamlessly aligns with the nature of inductive learning, which effectively generalizes from limited labeled data to larger test sets.

\subsection{Comparative Results}
\tab The effectiveness and robustness of the proposed embedding algorithm are tested by comparing its performances with all eight well-known embedding algorithms. The tests are done in two different categories: Binary classification and multi-class classification. It is observed that regardless of the complexity of the embedding algorithm, the efficiencies heavily depend on the number of classes that documents contain. SST-2, MR, and CoLA datasets are used for binary classification, while MR, SST-2, and CoLA datasets are for multi-class classification performance of the algorithms. Table~\ref{BinaryClassification} shows the results of the binary text classification on the three datasets, while the results of the multi-class text classification are reported in Table~\ref{MultiClassification}. The experimental results Tables~\ref{BinaryClassification} and~\ref{MultiClassification}, indicate the proposed model demonstrates superior performance in binary and multi-class text classification tasks, outperforming all baseline.

\begin{table}[H]
\centering
\caption{Comparative performance of our proposed method across binary text classification classifiers on MR, SST-2, and CoLA datasets, as evaluated by classification accuracy (Micro-F1 and Macro-F1). The highest scores are bold, and the second-highest scores are underlined.}
\label{BinaryClassification}
\resizebox{\textwidth}{!}{%
\begin{tabular}{ccccccc}
\hline
\multirow{2}{*}{\textbf{Models}} & \multicolumn{2}{c}{\textbf{MR}}       & \multicolumn{2}{c}{\textbf{SST-2}}    & \multicolumn{2}{c}{\textbf{CoLA}}     \\ \cline{2-7} 
                                 & Micro F1 & Macro F1 & Micro F1 & Macro F1 & Micro F1 & Macro F1 \\ \hline
\textbf{TF-IDF} $+$ \textbf{LR}             & 77.14             & 76.21             & 82                & 81.31             & 65.39             & 63.2              \\
\textbf{CNN-pretrain}            & 76.93             & 76.01             & 79.45             & 78.98             & 63.27             & 60.07             \\
\textbf{Bi-LSTM}                 & 76.81             & 76.21             & 79.6              & 79.4              & 63.21             & 60.1              \\
\textbf{SWEM}                    & 77.29             & 76.33             & 82.14             & 81.6              & 66.34             & 64.6              \\
\textbf{TextGCN}                & 77.43             & 77.02             & 82.45             & 82.4              & 68.1              & 66.71             \\
\textbf{TextING}                 & 79.69             & 79.1              & 83.11             & 83                & 69.4              & 67.11             \\
\textbf{BERT}                    & 86.6              & 85.31             & 91.53             & 91.02             & 82.17             & 77.08             \\
\textbf{VGCN-BERT}               & {\ul 86.73}       & {\ul 86.23}       & {\ul 91.94}       & {\ul 91.55}       & {\ul 83.91}       & {\ul 80.5}        \\
\textbf{GTPM}     & \textbf{87.41}    & \textbf{87.05}    & \textbf{92.45}    & \textbf{92.13}    & \textbf{85.14}    & \textbf{83.67}    \\ \hline
\end{tabular}%
}
\end{table}

\begin{table}[H]
\centering
\caption{Comparative performance of our proposed method across multi-class text classification classifiers on Reuters, 20NG, and Ohsumed datasets, as evaluated by classification accuracy (Micro-F1 and Macro-F1). The highest scores are bold, and the second-highest scores are underlined.}
\label{MultiClassification}
\resizebox{\textwidth}{!}{%
\begin{tabular}{ccccccc}
\hline
\multirow{2}{*}{\textbf{Models}} & \multicolumn{2}{c}{\textbf{Reuters}} & \multicolumn{2}{c}{\textbf{20NG}} & \multicolumn{2}{c}{\textbf{Ohsumed}} \\ \cline{2-7} 
                                 & Micro F1          & Macro F1         & Micro F1        & Macro F1        & Micro F1          & Macro F1         \\ \hline
\textbf{TF-IDF + LR}             & 84.75             & 84.52            & 79.75           & 79              & 60.42             & 52.2             \\
\textbf{CNN-pretrain}            & 86.66             & 86.41            & 81.09           & 80.18           & 62.45             & 57.13            \\
\textbf{Bi-LSTM}                 & 86.92             & 86.2             & 80.86           & 80.8            & 62.48             & 59.45            \\
\textbf{SWEM}                    & 87.13             & 86.86            & 81.77           & 81.69           & 68.03             & 63.2             \\
\textbf{TextGCN}                & 87.24             & 87.15            & {\ul 85.88}     & {\ul 85.34}     & 68.11             & 60.61            \\
\textbf{TextING}                 & 87.59             & 87.58            & 82.76           & 82.6            & 71.92             & 68.8             \\
\textbf{BERT}                    & 89.68             & 89.6             & 84.03           & 84.01           & 69.38             & 65.1             \\
\textbf{VGCN-BERT}               & {\ul 90.34}       & {\ul 90.3}       & 84.22           & 84.16           & {\ul 72.04}       & {\ul 68}         \\
\textbf{GTPM}     & \textbf{92.91}    & \textbf{92.8}    & \textbf{86.43}  & \textbf{86.09}  & \textbf{76.19}    & \textbf{71.11}   \\ \hline
\end{tabular}%
}
\end{table}

\section{Conclusions}
\tab A novel text embedding method, termed the Guided Transition Probability Matrix (GTPM) model, is introduced, focusing on leveraging the graph structure of sentences to construct embedding vectors. It aims to capture syntactic, semantic, and hidden content elements within text data. By employing random walks on a word graph generated from text, the model calculates transition probabilities to derive embedding vectors, effectively extracting semantic features. The method facilitates the extraction of language rules from textual documents, enhancing the understanding and representation of text data.

\tab We presented a comprehensive study on text classification utilizing graph-based embedding methods, mainly focusing on the proposed Guided Transition Probability Matrix (GTPM) approach. Our investigation encompassed various aspects of the text classification process, including feature extraction, parameter selection, robustness analysis, and comparative performance evaluation.

\tab The feature vectors obtained from the proposed GTPM embedding algorithm served as input for a multi-layer neural network architecture for text classification tasks. Through extensive experimentation, we demonstrated the effectiveness of the proposed method in achieving superior performance compared to baseline models across both binary and multi-class classification scenarios. The use of Keras API with TensorFlow backend facilitated the construction of a sequential neural network with appropriate activation functions and optimization techniques tailored to the specific classification tasks.

\tab Visual inspection aided by dimensional reduction techniques such as TSNE offers valuable insights into the performance and efficacy of classification algorithms. The distinct clustering of embedding vectors obtained from different methods on the Reuters dataset underscores the potential superiority of the proposed model in capturing meaningful features for classification tasks.

\tab Parameter selection played a crucial role in optimizing the performance of the embedding algorithm. Our experiments revealed that the performance of the GTPM approach was highly dependent on the number of walks per node and the length of walks. Optimal parameter values were determined through systematic experimentation, leading to improved classification accuracy.

\tab Moreover, the robustness of the proposed method was evaluated under constrained training data scenarios. By limiting the training set size to $10\%$ of the original dataset, we assessed the algorithm’s ability to generalize effectively to unseen data. The results demonstrated the superior robustness of the GTPM approach compared to baseline models, indicating its potential for real-world applications where labeled data may be limited.

\tab In comparative analyses, the proposed GTPM method consistently outperformed other state-of-the-art embedding algorithms across various datasets and classification tasks. The superior performance of our method underscores its efficacy in capturing meaningful features from text data, thereby enhancing the overall accuracy of text classification models.

\tab In conclusion, our study highlights the significance of graph-based embedding methods, particularly the GTPM approach, in advancing the field of text classification. The proposed method offers promising results in terms of both performance and robustness, paving the way for future research in text processing and natural language understanding.

\bibliographystyle{unsrt}
\bibliography{references}

\end{document}